\documentclass[10pt,twocolumn,letterpaper]{article}

\usepackage[pagenumbers]{cvpr} 

\newcommand{\ns}{$^{\circ}$}                 
\newcommand{\wmax}{w_{\max}}
\newcommand{\zinv}{z^{\mathrm{inv}}}
\newcommand{\zden}{z^{\mathrm{den}}}
\newcommand{\up}{$\uparrow$}
\newcommand{\dn}{$\downarrow$}
\newcommand{\fittable}[1]{\resizebox{\ifdim\width>\linewidth\linewidth\else\width\fi}{!}{#1}}  

\definecolor{cvprblue}{rgb}{0.21,0.49,0.74}
\usepackage[pagebackref,breaklinks,colorlinks,allcolors=cvprblue]{hyperref}

\def\confName{CVPR}
\def\confYear{2027}

\title{Fast Training-Free Low-Light Video Enhancement\\with a Frozen Image Diffusion Prior}

\author{Hon Ming Yam$^{1}$ \quad Chun Pong Lau$^{1}$\\
$^{1}$City University of Hong Kong \quad 
{\tt\small \{hmyam4, cplau27\}@cityu.edu.hk}
}

\begin{document}
\twocolumn[{%
\renewcommand\twocolumn[1][]{#1}%
\maketitle
\begin{center}
  \vspace{-1.5em}
  \includegraphics[width=0.8\linewidth]{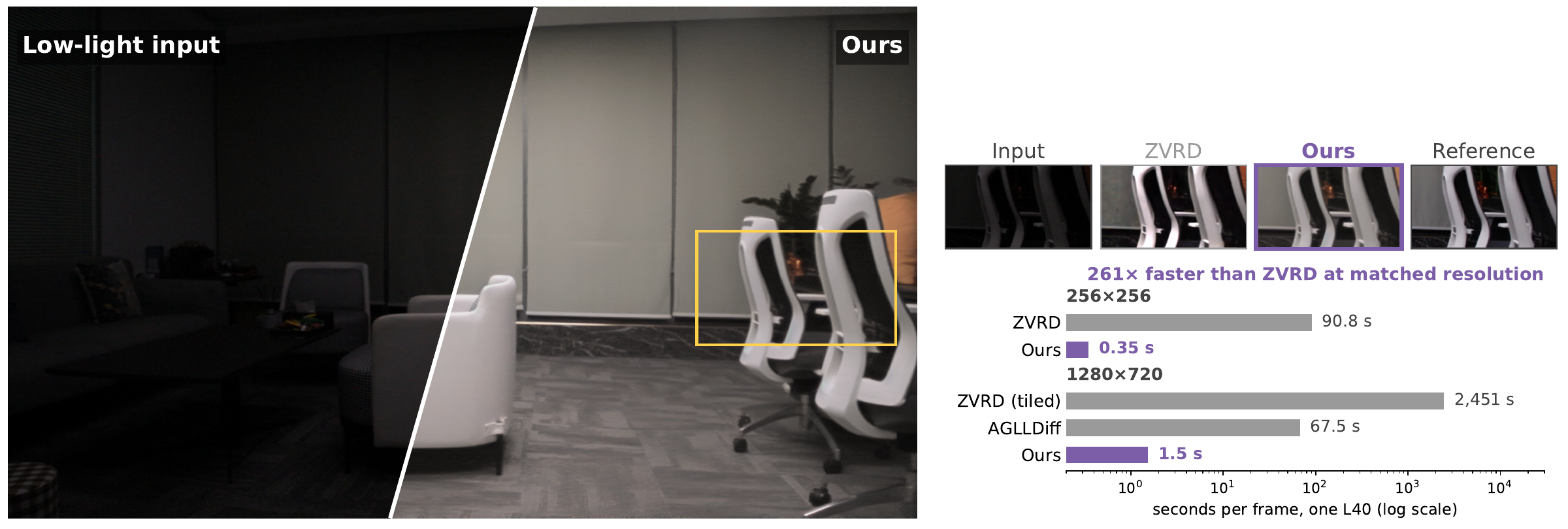}
  \captionof{figure}{\textbf{Training-free low-light video enhancement at a usable speed.}
  Left: a low-light SDSD frame, shown without brightening, and our enhancement.
  Right, top: the boxed region from the input, ZVRD (our implementation), our pipeline, and the reference; ZVRD leaves visible noise and color blotches.
  Right, bottom: seconds per frame on one NVIDIA L40 (log scale).
  At $256{\times}256$ the times are those of \cref{tab:zvrd}, where our pipeline is $261\times$ faster than ZVRD.
  At $1280{\times}720$, ZVRD tiles each frame and was timed on this clip only; AGLLDiff and our pipeline are averages over SDSD clips, and the gap to ZVRD is about $1{,}600\times$.
  The clip is one of the design clips of \cref{sec:data}.}
  \label{fig:teaser}
\end{center}%
}]
\begin{abstract}
Pretrained image diffusion models can enhance low-light video without low-light training data and without retraining for a new camera.
ZVRD, the only published training-free method for low-light video, is too slow to use: our implementation needs over a minute per $256{\times}256$ frame.
We present a causal, training-free pipeline that adds two temporal components to a frozen MR.~Illuminate image prior.
Temporal input fusion recursively averages motion-compensated low-light frames before diffusion inversion.
Late-start offset propagation runs the full sampling trajectory only on key frames.
Each non-key frame inverts the low-noise part of its own trajectory and adds an enhancement offset warped from the latest key frame.
At ZVRD's own resolution the pipeline is $261\times$ faster than ZVRD and scores better on all five metrics we report.
On 13 held-out SDSD clips at $1280{\times}720$, it improves on framewise eight-step processing by 0.55\,dB in PSNR and 0.034 in LPIPS and halves the warping error, in 10\% less time.
Measurements inside the model show why, and a linear-Gaussian model of inversion explains them.
Fusion removes frame-to-frame noise before inversion amplifies it.
A flow error perturbs the transported offset ten times less than it perturbs a transported state.
The perceptual gain does not transfer to low resolution: at $346{\times}260$, LPIPS is worse by 0.028.
At 0.65 frames per second, the pipeline is an offline method.
\end{abstract}

\section{Introduction}
\label{sec:intro}

Low-light video enhancement must improve visibility, keep detail, and avoid distracting changes between frames.
Learned video enhancers do this after training on low-light video~\cite{wang2021sdsd,fu2023did,lin2024bvirlv}.
Such data are expensive to capture, and supporting a new camera or condition can call for new data and retraining.
Pretrained image diffusion models offer another route.
MR.~Illuminate~\cite{cho2026mr}, hereafter MR, enhances a low-light image with a frozen latent diffusion model through inversion, attention injection, and decoding, with no training or test-time optimization.

Extending such priors to video has been costly.
To our knowledge, ZVRD~\cite{cao2025zvrd} is the only published training-free method for low-light video enhancement.
It adds temporal attention, temporal guidance, and noise sharing to a pretrained image diffusion model, samples with 950 steps, and optimizes enhancement parameters at test time.
ZVRD has no public code.
Our implementation takes 90.8\,s per frame at the published $256{\times}256$ geometry on one NVIDIA L40 (\cref{tab:zvrd}), and about 2{,}450\,s at $1280{\times}720$, where each frame is tiled into 50 patches (\cref{fig:teaser}).
Its temporal components also do not make it stable: at matched resolution its flicker is $2.7\times$ and its LPIPS almost twice that of framewise MR with eight steps, and it leaves visible noise and color blotches (\cref{tab:zvrd,fig:teaser}).

We ask what a practical training-free diffusion pipeline for low-light video requires.
The starting point is simple: framewise MR with eight sampling steps is already $239\times$ faster than ZVRD at matched resolution and scores better on every metric (\cref{sec:zvrd}).
Two observations then guide the temporal design.

First, sensor noise is the dominant per-frame degradation in low-light video.
MR inverts each frame into its diffusion state, so each frame's noise realization enters its inversion latent, its attention features, and its decoder skips.
We therefore apply a classical remedy, motion-compensated recursive temporal filtering~\cite{dubois1984noise}, to the input of the diffusion model.

Second, the enhancement that a full diffusion trajectory applies changes slowly between adjacent frames, while the trajectory is expensive.
SDEdit~\cite{meng2022sdedit} starts denoising from an intermediate noise level; we use this idea across frames.
A key frame runs the full trajectory and records how far its denoised state at an intermediate time departs from its own inversion at that time.
Each non-key frame inverts only to that time, adds the warped departure, and denoises from there with its own encoding, attention features, and decoder skips.

The two components are complementary: alone, late-start propagation raises warping error and input fusion adds time; together they are faster and better than framewise processing (\cref{tab:ablation}).
Our contributions are:
\begin{itemize}
  \item A causal, training-free pipeline for low-light video that needs no training data, retraining, or test-time optimization. At ZVRD's own $256{\times}256$ resolution it is $261\times$ faster than ZVRD and better on all five metrics we report (\cref{sec:zvrd}).
  \item Two temporal components for a frozen diffusion prior: \emph{temporal input fusion}, which filters sensor noise before inversion, and \emph{late-start offset propagation}, which starts non-key frames from an intermediate time with a transported offset. On held-out $1280{\times}720$ clips they improve peak signal-to-noise ratio (PSNR), learned perceptual image patch similarity (LPIPS), and warping error over framewise processing in 10\% less time, and the gain holds over 300-frame sequences (\cref{sec:heldout,sec:transfer}).
  \item An analysis of why the components work together, with measurements inside the model and a linear-Gaussian model of inversion, and of where they fail: the perceptual gain is lost at low resolution (\cref{sec:inside,sec:transfer}).
\end{itemize}

\section{Related Work}
\label{sec:related}

\paragraph{Low-light enhancement and training data.}
Low-light image enhancement ranges from Retinex-based estimation without learning~\cite{guo2017lime} to models trained on paired images~\cite{wei2018retinexnet,yang2021lolv2,cai2023retinexformer}, on unpaired images~\cite{jiang2021enlightengan}, or without references but still on low-light data~\cite{guo2020zerodce,ma2022sci}.
Low-light video enhancement has mostly followed the learned route, with paired videos from dedicated rigs~\cite{jiang2019smoid,chen2019smid,wang2021sdsd}, synthesized from images~\cite{zhang2021stablellve}, or collected in larger benchmarks~\cite{fu2023did,lin2024bvirlv}.
Zero-TIG~\cite{li2025zerotig} needs no paired references but trains its network on the BVI-RLV video dataset.
ReDDiT~\cite{lan2026reddit} distills a diffusion enhancer into a few-step student trained on LOLv2 pairs.
Among training-free image methods, MR~\cite{cho2026mr} is our backbone, FourierDiff~\cite{lv2024fourierdiff} embeds Fourier priors in a pretrained diffusion model, and AGLLDiff~\cite{lin2024aglldiff} guides sampling with attribute gradients at inference.
Image enhancers applied frame by frame are unstable over time: on held-out SDSD clips the three ReDDiT models have $3.8$--$4.8\times$ our warping error (\cref{tab:heldout}).
Zero-TIG's flow feedback helps, but its warping error is still twice ours and its LPIPS higher.
AGLLDiff's guidance gradients cost 67\,s per $1280{\times}720$ frame, and it has the lowest PSNR and highest LPIPS of all methods we ran.

\paragraph{Training-free diffusion video restoration and editing.}
Zero-shot restoration with diffusion priors solves inverse problems with known degradation operators~\cite{kawar2022ddrm,rout2023psld,fei2023gdp}.
ZVRD~\cite{cao2025zvrd} extends it to video with temporal attention, temporal guidance, and noise sharing, and its follow-up adds video diffusion models~\cite{cao2026zvrv}.
Zero-shot video generation and editing methods share attention or features across frames~\cite{khachatryan2023t2vzero,yang2023rerender}.
TokenFlow~\cite{geyer2023tokenflow} propagates diffusion features along inter-frame correspondences, FRESCO~\cite{yang2024fresco} adds intra-frame correspondence and feature optimization, RAVE~\cite{kara2024rave} shuffles frames between grids during denoising, and DiffIR2VR-Zero~\cite{yeh2025diffir} adapts restoration diffusion models to video with latent warping and token merging.
These methods carry attention features or latent states from frame to frame.
For low-light enhancement, a baseline in this spirit, which warps the key frame's inversion state and attention bank to the other frames, is slower than framewise MR and worse on every metric (\cref{tab:heldout}): the transported state carries the key frame's noise.
Late-start propagation differs in what it transports: one latent offset at an intermediate time, added to the current frame's own partial inversion, while attention features always come from the current frame.

\paragraph{Diffusion acceleration and feature reuse.}
Denoising diffusion models~\cite{ho2020ddpm} originally needed hundreds of sampling steps; DDIM~\cite{song2022ddim} and higher-order solvers~\cite{lu2022dpmsolver} trade steps against quality.
Fewer steps alone leave each frame's noise in place: with 8, 15, or 25 steps, framewise MR has $1.9$--$2.4\times$ our warping error (\cref{tab:heldout}).
Deterministic DDIM inversion maps an image to a noise latent; editing methods refine it~\cite{mokady2023nulltext} or inject features captured during it~\cite{tumanyan2023pnp}, as MR does.
SDEdit~\cite{meng2022sdedit} starts generation from an intermediate noise level.
Late-start propagation is closest to it, but its starting state combines the frame's own inversion with an offset from a key frame.
DeepCache~\cite{ma2024deepcache} and Clockwork Diffusion~\cite{habibian2024clockwork} reuse U-Net features across the steps of one image and are complementary to reuse across frames.

\paragraph{Temporal filtering of noisy video.}
Motion-compensated recursive filtering averages each frame with the motion-compensated previous output and lowers the weight where compensation fails~\cite{dubois1984noise}.
Learned video denoisers~\cite{tassano2020fastdvdnet} require training.
We apply the classical filter to the input of a training-free diffusion prior.

\begin{figure*}[t]
  \centering
  \includegraphics[width=0.92\linewidth]{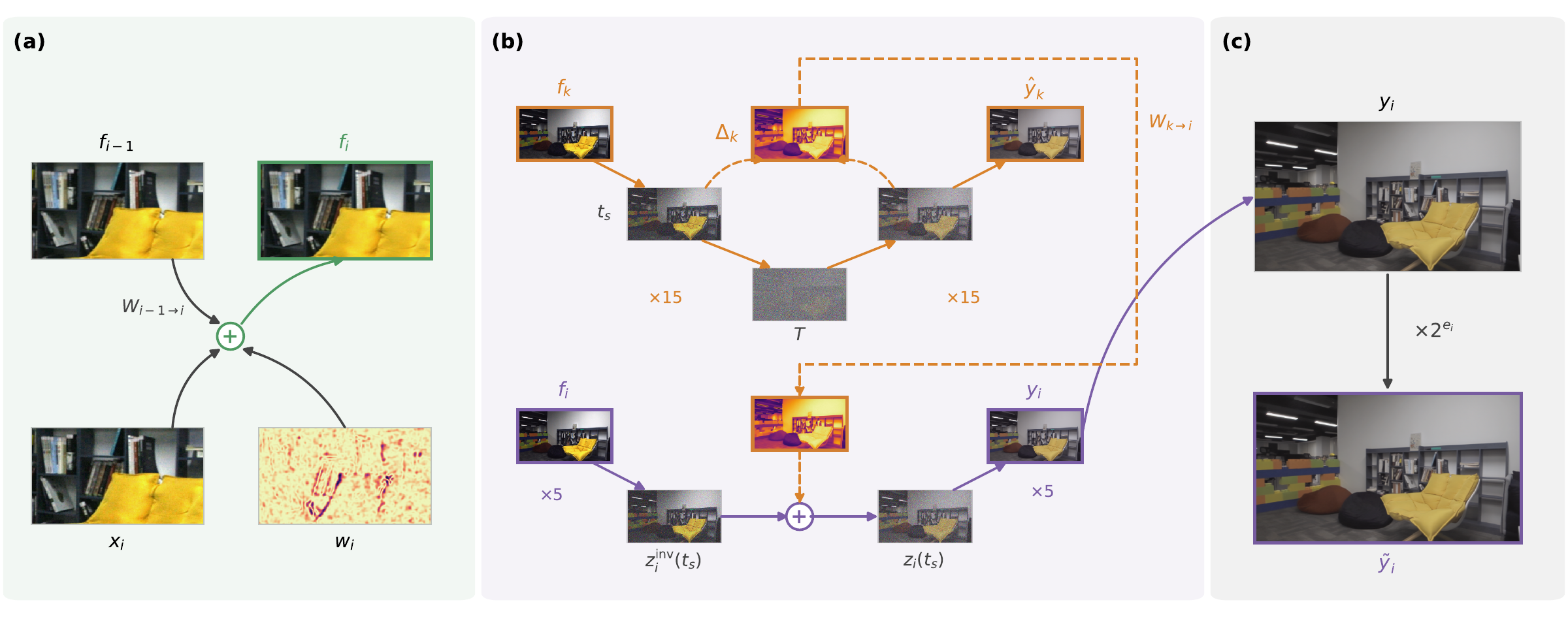}
  \caption{\textbf{Pipeline overview} on real SDSD frames (low-light images shown with a display gain).
  (a)~Temporal input fusion averages the current frame $x_i$ with the flow-warped previous fused input, using a per-pixel weight $w_i$ (dark: flow unreliable or content changed).
  (b)~A key frame (orange) runs the full trajectory and stores the latent offset $\Delta_k$ at time $t_s$; a non-key frame (purple) inverts only to $t_s$, adds the warped offset, and denoises from there. Trajectory thumbnails are illustrative, and $\Delta_k$ is drawn in image space for display.
  (c)~A causal exposure gain gives the output. Key frames cost 30 U-Net evaluations and non-key frames 10.}
  \label{fig:overview}
\end{figure*}

\section{Method}
\label{sec:method}

\subsection{Setting and overview}
\label{sec:setting}

Let $x_i$ be the $i$-th low-light RGB frame, $y_i$ its enhanced output, and $\tilde y_i$ the output after exposure calibration.
Inference is causal: it uses $x_0,\dots,x_i$ and never a future frame or a normal-light reference.
By \emph{training-free} we mean that no network weight is trained or fine-tuned for low-light enhancement, no low-light training data are collected, and no gradient-based optimization runs at test time.
The pipeline has a small number of hand-set constants; \cref{sec:data} states which clips were used to choose them.

\Cref{fig:overview} shows the four parts: a frozen MR backbone, temporal input fusion, late-start offset propagation, and a causal exposure gain.

\subsection{Framewise MR backbone}
\label{sec:backbone}

Each frame is encoded by the variational autoencoder (VAE) of Stable Diffusion v1.5~\cite{rombach2022ldm}; its encoder skip features are kept for decoding.
Deterministic inversion with denoising diffusion implicit models (DDIM)~\cite{song2022ddim} maps the latent towards noise with the denoising U-Net and records, at each inversion time, the self-attention outputs of the nine up-block layers.
We call this record the attention bank.
The inversion endpoint is normalized to the channel statistics of a fixed Gaussian noise sample by adaptive instance normalization (AdaIN)~\cite{huang2017adain}.
Denoising then injects the attention bank, and the final latent is decoded with the frame's own skips~\cite{cho2026mr}.

A schedule is a set of diffusion times used for inversion and, reversed, for denoising.
MR's default schedule, A25, has 25 times and costs 50 U-Net evaluations per frame.
We use two shorter framewise schedules as controls: E15 with $\mathcal T_{15}=\{1,67,133,\dots,925\}$ (30 evaluations) and E8 with $\mathcal T_{8}=\{1,126,\dots,876\}$ (16 evaluations).
We write NFE for the number of network evaluations per frame.
Implementation constants are in Supp.~\cref{sec:supp-impl}.

\subsection{Temporal input fusion}
\label{sec:fusion}

The model input $f_i$ is a causal recursive average of the low-light frames, with $f_0=x_0$ and
\begin{align}
  f_i &= (1-w_i)\odot x_i+w_i\odot \hat f_i, \label{eq:fusion}\\
  \hat f_i &= W_{i-1\to i}(f_{i-1}). \label{eq:history}
\end{align}
Here $W_{i-1\to i}$ is a backward bilinear warp along Farneb\"ack flow~\cite{farneback2003} between $x_{i-1}$ and $x_i$, so $\hat f_i$ is the motion-compensated history.
Let $R_i\in\{0,1\}$ mark pixels that pass a forward-backward consistency check.
The per-pixel weight is
\begin{align}
  w_i &= \wmax\,R_i\,\exp\!\Bigl(-\bigl(d_i/(2\hat\sigma_i+10^{-3})\bigr)^2\Bigr), \label{eq:weight}\\
  d_i &= \bigl|\,G_2*\bigl(Y(x_i)-Y(\hat f_i)\bigr)\bigr|, \label{eq:gate}
\end{align}
where $Y$ is the Rec.\,709 luma of sRGB values in $[0,1]$, $G_s$ is a Gaussian blur with standard deviation $s$ pixels, and $\hat\sigma_i=1.4826\,\mathrm{median}\,|Y(x_i)-G_{1.5}*Y(x_i)|$ is a robust high-frequency scale of the frame.
The exponential term lowers the weight where the warped history disagrees with the current frame; at a difference of twice this scale the weight has fallen to $\wmax/e$.
We use $\wmax=0.8$, an effective memory of about five frames.
Fusion changes only the input of the diffusion model; calibration and evaluation use the original $x_i$.

\subsection{Late-start offset propagation}
\label{sec:latestart}

Frames with $i\equiv 0 \pmod K$, $K=10$, are key frames.
A key frame runs the full E15 trajectory on its fused input, up to the endpoint $T=925$, and records two latents at an intermediate time $t_s=265$: the inversion latent $\zinv_k(t_s)$ and the denoising latent $\zden_k(t_s)$.
Their difference
\begin{equation}
  \Delta_k=\zden_k(t_s)-\zinv_k(t_s)
  \label{eq:offset}
\end{equation}
is the enhancement that the full trajectory has applied by the time it returns to $t_s$.
We store $\Delta_k$, its per-channel spatial mean $\bar\Delta_k$, and the key frame's fused input.

A non-key frame $i$ encodes its fused input and inverts only over $\mathcal T_s=\{t\in\mathcal T_{15}: t\le t_s\}=\{1,67,133,199,265\}$, recording its own attention bank.
Flow from the key frame's fused input to the current one is resized to the latent grid; $W_{k\to i}$ is bilinear sampling on this grid, and $M_i=1$ marks latent locations with reliable flow over at least 99.9\% of their image footprint.
The starting state is
\begin{equation}
\begin{aligned}
  z_i(t_s) = \zinv_i(t_s) &+ M_i\odot W_{k\to i}(\Delta_k)\\
                          &+ (1-M_i)\odot\bar\Delta_k .
\end{aligned}
  \label{eq:start}
\end{equation}
Denoising runs over $\mathcal T_s$ in reverse with the frame's own attention bank, without AdaIN, and the result is decoded with the frame's own skips.

Only one latent offset is transported per key interval.
Attention features and enhanced images are never transported, and the offset is replaced at the next key frame, so errors do not accumulate.
A non-key frame costs 10 U-Net evaluations and a key frame 30.
A sequence of $N$ frames therefore needs $10+20\lceil N/K\rceil/N$ evaluations per frame on average: 12 for $N=30$ or $300$, against 16 for E8 and 30 for E15.

\subsection{Exposure calibration}
\label{sec:calibration}

A causal post-process corrects global brightness with one scalar gain $2^{e_i}$ applied to linear RGB.
Let $\mu_x$ and $\mu_y$ be the mean luminances of $x_i$ and $y_i$, and $q_c$ the 0.995 luminance quantile of $x_i$.
The requested correction is
\begin{equation}
  r_i=\tfrac{1}{2}\log_2\frac{0.9\,\mu_x/q_c}{\mu_y},
  \label{eq:calib}
\end{equation}
clipped to $[-0.5,0.5]$ exposure values.
A guard cancels a brightening request when the 0.999 quantile would no longer call for it, and the gain is smoothed over time as $e_i=0.8\,e_{i-1}+0.2\,\hat e_i$, where $\hat e_i$ is the guarded request (Supp.~\cref{sec:supp-calib}).
The calibrator never sees a reference.
It is applied identically to every MR configuration in this paper, and we do not claim it as a contribution.

\section{Experimental Setup}
\label{sec:setup}

\subsection{Data and splits}
\label{sec:data}

\paragraph{SDSD.}
SDSD~\cite{wang2021sdsd} provides paired low- and normal-light videos of dynamic scenes.
We use its 25 test scenes (12 indoor, 13 outdoor), the first 30 frames of each, resized from $1920{\times}1080$ to $1280{\times}720$.
We split them into 12 \emph{design} clips and 13 \emph{held-out} clips (Supp.~\cref{tab:supp-splits}).
The settings of the two temporal components were chosen on the design clips only: $\wmax$ and $t_s$ each from two candidate values, and $K$ from $\{5,10,20\}$ for the state-reuse baseline of \cref{sec:baselines}.
The gate constants of \cref{eq:weight,eq:gate} were set by hand and not varied.
All SDSD results at $1280{\times}720$ are reported on the held-out clips.
The E8 schedule and the calibration constants were also set on SDSD: the calibration used six of the held-out clips, and E8 was chosen from framewise measurements on the SDSD test scenes.
We therefore also report the seven held-out clips that informed neither the temporal components nor the calibration.

\paragraph{SDE.}
The independent dataset is the RGB part of SDE~\cite{liang2024evlight}: the first 30 frames of its 15 test scenes at the native $346{\times}260$, without events.
No setting was chosen on SDE.
Our full-frame RGB protocol is not comparable with published event-guided results.

\paragraph{Long sequences and stress tests.}
The long-sequence set contains the first 300 frames of four outdoor SDSD scenes that are outside both splits.
Two stress sets reuse these windows, subsampled with stride 2 and 4 (30 frames each), to simulate faster motion.
A third joins the first five frames of one held-out clip to the first 25 frames of another, six times, so that a scene cut falls inside a key interval.

We did not evaluate on DID~\cite{fu2023did}, which we could not obtain, or on BVI-RLV~\cite{lin2024bvirlv}, Zero-TIG's training set.

\subsection{Compared methods}
\label{sec:baselines}

The only training-free method for low-light video is ZVRD, which we compare at its own resolution (\cref{sec:zvrd}).
At $1280{\times}720$ the training-free baselines are AGLLDiff~\cite{lin2024aglldiff}, run framewise, and MR with its default A25 schedule and with E15 and E8.
We add a \emph{state-reuse} baseline on the same backbone, in the spirit of feature propagation~\cite{geyer2023tokenflow}: non-key frames warp the key frame's inversion endpoint and attention bank and run all 15 denoising steps (16.5 NFE).
Zero-TIG~\cite{li2025zerotig} and ReDDiT with 2, 4, and 8 steps~\cite{lan2026reddit} use low-light training data; we run their public checkpoints as references.
DiffIR2VR-Zero, TokenFlow, and VISION-XL~\cite{yeh2025diffir,geyer2023tokenflow,kwon2025visionxl} have no low-light configuration, and adapting them to MR's backbone would be a new method, so we do not include them.
We did not run FourierDiff~\cite{lv2024fourierdiff}; AGLLDiff represents guided training-free image diffusion.

\subsection{Metrics and timing}
\label{sec:metrics}

We report peak signal-to-noise ratio (PSNR) on full RGB, the structural similarity index (SSIM)~\cite{wang2004ssim}, and the learned perceptual image patch similarity (LPIPS)~\cite{zhang2018lpips} with AlexNet.
No reference-derived brightness correction enters any score.
The two temporal metrics use Farneb\"ack flow estimated on the normal-light references, with one validity mask shared by all methods.
\emph{Warping error}~\cite{lai2018blind} is the mean absolute RGB difference between the current output and the warped previous output.
\emph{Flicker} is the mean absolute frame-to-frame change, along the reference flow, of the log-luminance ratio between output and reference, so brightness changes that the reference also shows are not counted.

Clips are weighted equally.
For a difference between two methods we give a 95\% interval from a paired bootstrap over clips (10{,}000 draws); on sets of four or six sequences the intervals are indicative only.
All tables share one format: bold marks the best value among training-free methods, and $^{\circ}$ marks a value whose difference from our pipeline has an interval that includes zero.

Times are seconds per frame on one NVIDIA L40 with batch size one, from image decoding to the final output, including flow, fusion, and calibration.
Where we state that a time is relative to E8 in the same process, E8 was re-timed in the process that timed the method; all other times are separate single runs.

\section{Results}
\label{sec:results}

\subsection{Comparison with ZVRD at its own resolution}
\label{sec:zvrd}

\begin{table}[t]
  \centering
  \setlength{\tabcolsep}{3pt}
  \fittable{%
  \begin{tabular}{@{}lccccccrr@{}}
    \toprule
    Method & PSNR\up & SSIM\up & LPIPS\dn & Warp\dn & Flicker\dn & NFE & s/frame\dn & Speedup \\
    \midrule
    ZVRD~\cite{cao2025zvrd} (our impl.) & 15.81\ns & 0.512 & 0.312 & 1.51 & 0.280 & 950 & 90.78 & 1$\times$ \\
    MR E15 & 19.08\ns & 0.681\ns & \textbf{0.156} & 1.00 & 0.107 & 30 & 0.67 & 135$\times$ \\
    MR E8 & 18.91\ns & 0.672 & 0.163 & 0.94 & 0.103 & 16 & 0.38 & 239$\times$ \\
    Ours & \textbf{19.18} & \textbf{0.688} & 0.178 & \textbf{0.51} & \textbf{0.087} & \textbf{12} & \textbf{0.35} & \textbf{261$\times$} \\
    \bottomrule
  \end{tabular}}
  \caption{\textbf{Comparison with ZVRD at $256{\times}256$}, six SDSD clips of 30 frames, all methods training-free.
  MR E15 and MR E8 are framewise MR with 15 and 8 sampling steps.
  Warp is warping error in units of $10^{-2}$; NFE is the number of network evaluations per frame; s/frame is measured on one NVIDIA L40.
  Speedup is the geometric mean over clips of the time ratio to ZVRD.
  Bold: best in column. $^{\circ}$: the 95\% paired-bootstrap interval of the difference from our pipeline includes zero.
  The six clips are design clips (\cref{sec:data}); no setting was chosen at this resolution.}
  \label{tab:zvrd}
\end{table}

At $1280{\times}720$ ZVRD's cost is dominated by tiling, so we compare at its published $256{\times}256$ geometry.
Six SDSD clips are center-cropped and resized to $256{\times}256$.
ZVRD runs with its 950-step configuration and RAFT flow~\cite{teed2020raft} (Supp.~\cref{sec:supp-zvrd}).

\Cref{tab:zvrd} shows that framewise E8 already scores better than ZVRD on all five metrics, at $239\times$ lower cost.
Our pipeline is $261\times$ faster than ZVRD and also better on all five metrics; it is better in PSNR and in LPIPS on five of the six clips, and the PSNR interval includes zero.
Compared with E8, the pipeline lowers warping error by 46\% and flicker by 16\% at this resolution, but its LPIPS is worse by 0.015.
\Cref{sec:transfer} analyzes this loss.

\subsection{Held-out SDSD clips at \texorpdfstring{$1280{\times}720$}{1280x720}}
\label{sec:heldout}

\begin{table*}[t]
  \centering
  \setlength{\tabcolsep}{5pt}
  \fittable{%
  \begin{tabular}{@{}llccccccr@{}}
    \toprule
    Method & Low-light training data & PSNR\up & SSIM\up & LPIPS\dn & Warp\dn & Flicker\dn & NFE & s/frame\dn \\
    \midrule
    ReDDiT-2~\cite{lan2026reddit} & LOLv2 pairs & 17.25\ns & 0.617 & 0.286 & 1.99 & 0.166 & 2 & 0.47 \\
    ReDDiT-4~\cite{lan2026reddit} & LOLv2 pairs & 18.04\ns & 0.582 & 0.302 & 2.48 & 0.184 & 4 & 0.87 \\
    ReDDiT-8~\cite{lan2026reddit} & LOLv2 pairs & 18.51\ns & 0.609 & 0.295 & 2.33 & 0.187 & 8 & 1.67 \\
    Zero-TIG~\cite{li2025zerotig} & BVI-RLV videos & 19.09\ns & 0.739 & 0.296 & 1.10 & 0.131 & n/a & 0.12 \\
    \midrule
    AGLLDiff~\cite{lin2024aglldiff} & none & 13.35 & 0.589 & 0.378 & 1.39 & 0.134 & 49.9 & 67.45 \\
    MR A25~\cite{cho2026mr} & none & 17.47 & 0.672 & 0.233\ns & 1.23 & 0.136 & 50 & 3.81 \\
    MR E15 & none & 17.35 & 0.673 & 0.235\ns & 1.15 & 0.132 & 30 & 2.62 \\
    MR E8 & none & 17.21 & 0.674 & 0.252 & 1.01 & 0.126 & 16 & 1.70 \\
    MR state reuse & none & 16.99 & 0.665 & 0.281 & 1.21 & 0.137 & 16.5 & 1.87 \\
    Ours & none & \textbf{17.76} & \textbf{0.710} & \textbf{0.218} & \textbf{0.52} & \textbf{0.109} & \textbf{12} & \textbf{1.52} \\
    \bottomrule
  \end{tabular}}
  \caption{\textbf{Held-out SDSD clips}, 13 clips of 30 frames at $1280{\times}720$.
  The first four rows use low-light training data and are references; they are not ranked.
  MR A25, E15, and E8 are framewise MR with 25, 15, and 8 sampling steps; MR state reuse warps the key frame's inversion state to the other frames (\cref{sec:baselines}).
  Warp is warping error in units of $10^{-2}$. NFE is the number of network evaluations per frame, averaged over frames (Zero-TIG is not a diffusion model).
  Bold: best among training-free methods. $^{\circ}$: the 95\% paired-bootstrap interval of the difference from our pipeline includes zero.
  All MR rows and ours include the exposure calibration of \cref{sec:calibration}; the external methods run as released.
  Times are single runs on one NVIDIA L40 and indicate magnitude only, except that E8 was timed in the same process as our pipeline. The E15 run does not skip replaced attention layers; with the skip E15 takes 2.44\,s (Supp.~\cref{tab:supp-runtime}).}
  \label{tab:heldout}
\end{table*}

\begin{figure*}[t]
  \centering
  \includegraphics[width=0.86\linewidth]{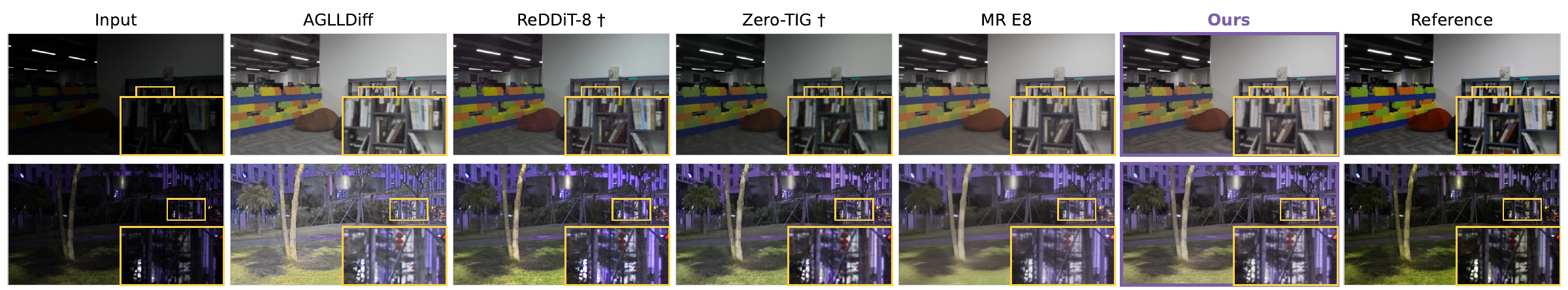}
  \caption{\textbf{Qualitative comparison} on frame 15, a non-key frame, of the held-out indoor and outdoor clips whose LPIPS change relative to E8 is the median of their group.
  Inputs are shown without brightening; insets enlarge the most textured window of the reference (yellow box).
  $\dagger$ marks methods that require training.
  Our pipeline shows less residual grain than framewise MR E8 on the shelves and walls; the trained references render brighter images with higher contrast.}
  \label{fig:qualitative}
\end{figure*}

\paragraph{Against framewise processing.}
On the 13 held-out clips (\cref{tab:heldout}), the pipeline improves on E8 by 0.55\,dB in PSNR (95\% interval 0.27--0.79) and by 0.034 in LPIPS (0.018--0.050).
It lowers warping error by 49\% and flicker by 14\%, and takes $0.90\times$ the time of E8 run in the same process.
PSNR is higher on 11 clips, LPIPS lower on 12, and both temporal metrics are lower on all 13.
On the seven clips that informed neither the temporal components nor the calibration, the LPIPS gain (0.027, interval 0.016--0.039) and the temporal gains hold on every clip; the PSNR gain is 0.38\,dB and its interval includes zero.

Longer schedules do not close the gap.
Against E15 and against MR's default A25, the pipeline gains 0.41 and 0.30\,dB in PSNR and lowers warping error by 55\% and 58\%, in roughly 60\% and 40\% of their time.
Its LPIPS is lower than theirs by 0.017 and 0.015, but these two differences are within the bootstrap interval.

\paragraph{Against other methods.}
Among training-free methods the pipeline is best on all five metrics.
The trained Zero-TIG has higher PSNR, by 1.33\,dB, although this interval narrowly includes zero; it also has higher SSIM and is about $12\times$ faster.
Our pipeline has lower LPIPS, warping error, and flicker than Zero-TIG on all 13 clips.

Calibration matters for these PSNR comparisons.
It accounts for 1.52\,dB of the pipeline's PSNR (\cref{tab:ablation}), and the external models have no comparable step.
Without calibration the pipeline's PSNR (16.24\,dB) is below that of ReDDiT and Zero-TIG, while its LPIPS (0.228) remains lower than that of every external method and of calibrated E8.
The perceptual and temporal gains thus do not depend on calibration; the PSNR ranking against external methods does.
\Cref{fig:qualitative} shows the lower residual grain on typical frames.

\subsection{Ablation}
\label{sec:ablation}

\begin{table}[t]
  \centering
  \setlength{\tabcolsep}{3pt}
  \fittable{%
  \begin{tabular}{@{}cllcccccc@{}}
    \toprule
    $\wmax$ & Trajectory & Calib. & PSNR\up & SSIM\up & LPIPS\dn & Warp\dn & Flicker\dn & Time\dn \\
    \midrule
    -- & E8 & yes & 17.21 & 0.674 & 0.252 & 1.01 & 0.126 & 1.00 \\
    -- & E15 & yes & 17.35 & 0.673 & 0.235\ns & 1.15 & 0.132 & 1.45$^{\ddagger}$ \\
    0.8 & E8 & yes & 17.49\ns & 0.707\ns & 0.228 & \textbf{0.50} & \textbf{0.109} & 1.03 \\
    0.8 & E15 & yes & 17.71\ns & \textbf{0.710}\ns & 0.213 & 0.54 & 0.110 & 1.49$^{\ast}$ \\
    -- & late start, offset & yes & 17.34 & 0.673 & 0.235\ns & 1.12 & 0.131 & \textbf{0.87} \\
    0.8 & late start, key state & yes & 17.57 & 0.696 & 0.249 & 0.56 & 0.112 & 0.91$^{\ast}$ \\
    0.5 & late start, offset & yes & 17.56 & 0.698 & \textbf{0.206} & 0.69 & 0.114 & 0.90 \\
    0.8 & late start, offset & no & 16.24 & 0.674 & 0.228 & 0.57 & \textbf{0.109} & 0.89 \\
    0.8 & late start, offset (ours) & yes & \textbf{17.76} & \textbf{0.710} & 0.218 & 0.52 & \textbf{0.109} & 0.90 \\
    \bottomrule
  \end{tabular}}
  \caption{\textbf{Ablation on the 13 held-out clips} at $1280{\times}720$.
  $\wmax$: fusion ceiling (-- means no fusion). ``Key state'' starts non-key frames from the warped key-frame state instead of \cref{eq:start}.
  E8 and E15 are framewise schedules with 8 and 15 steps; late start is \cref{sec:latestart}. Calib.: with exposure calibration.
  Warp is in units of $10^{-2}$. Time is relative to E8 timed in the same process, except $^{\ast}$, a ratio of separate single runs, and $^{\ddagger}$, from the repeated timing of Supp.~\cref{tab:supp-runtime}.
  Bold: best in column. $^{\circ}$: the 95\% paired-bootstrap interval of the difference from our pipeline (last row) includes zero.}
  \label{tab:ablation}
\end{table}

\begin{figure*}[t]
  \centering
  \includegraphics[width=0.84\linewidth]{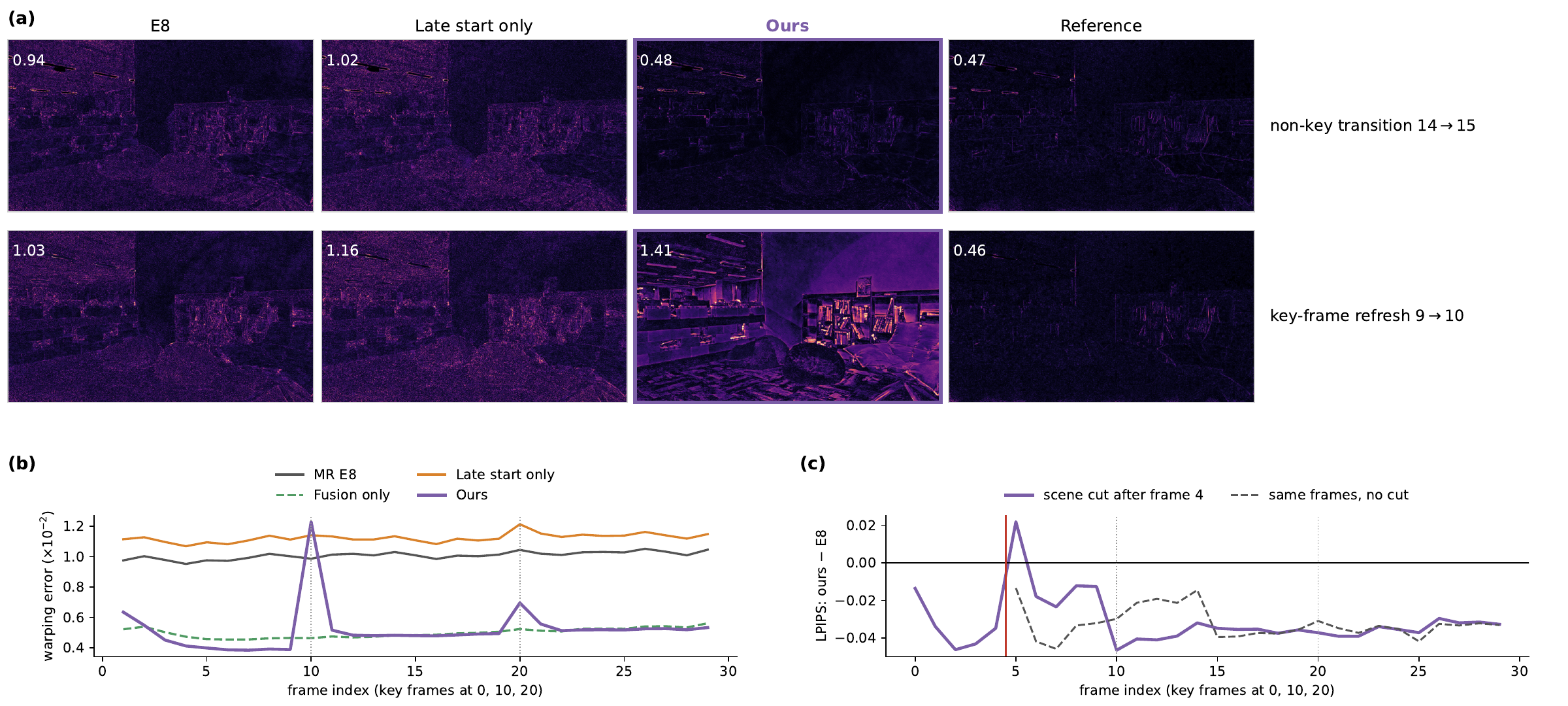}
  \caption{\textbf{Temporal behavior.}
  (a)~Temporal residual $|y_i-W(y_{i-1})|$ on a held-out clip, warped with the reference flow, at a non-key transition (top) and at the first key-frame refresh (bottom), for E8, late start without fusion, our pipeline, and the reference; numbers are mean residuals in units of $10^{-2}$.
  (b)~Warping error per frame index, averaged over the 13 held-out clips, for E8, each component alone, and our pipeline; dotted lines mark key frames.
  (c)~LPIPS of our pipeline minus E8 per frame index on the six scene-cut sequences (solid) and on the same frames without a cut (dashed); the red line marks the cut.}
  \label{fig:temporal}
\end{figure*}

\paragraph{Each component alone fails one requirement.}
\Cref{tab:ablation} shows that fusion alone improves every quality metric but is slower than E8 ($1.03\times$).
Late-start propagation alone is faster ($0.87\times$) but raises warping error by 11\%, on 12 of the 13 clips.
Combined, they are faster than E8 and better on every metric.

\paragraph{Late start is an accelerator.}
Fusion on the full E15 trajectory is the quality reference for our pipeline.
The pipeline matches its PSNR (the difference of 0.05\,dB is within the interval) and loses 0.005 in LPIPS, in 61\% of its time.

\paragraph{The transported quantity must be the offset.}
If non-key frames start from the warped key-frame state $\zden_k(t_s)$, the LPIPS gain over E8 disappears (0.003, within the interval) at the same cost.
Our pipeline has lower LPIPS than this variant on all 13 clips, by 0.031.
\Cref{sec:inside} explains why.

\paragraph{Fusion weight.}
A lower ceiling, $\wmax=0.5$, gives the best LPIPS in \cref{tab:ablation} but keeps less of the temporal gain: warping error falls by 31\% relative to E8 instead of 49\%, and PSNR rises by 0.35\,dB instead of 0.55.

\paragraph{Where the temporal gain occurs.}
\Cref{fig:temporal}b shows warping error per frame.
On non-key transitions it is about half that of E8 ($0.49$ against $1.01$, in units of $10^{-2}$).
At key-frame refreshes it jumps.
At the first refresh, frame 10, it exceeds E8 ($1.23$ against $0.99$, on 10 of the 13 clips); at the second it stays below ($0.70$ against $1.04$).
We attribute this to start-up: the first key frame is unfused, $f_0=x_0$, so the fused key frame at frame 10 changes the appearance most (\cref{fig:temporal}a).

\subsection{Why the components work together}
\label{sec:inside}

Two measurements inside the model, made on the six design clips of \cref{tab:zvrd} at $1280{\times}720$, explain the ablation (Supp.~\cref{sec:supp-mechanism}).

\paragraph{Inversion amplifies frame-to-frame noise.}
We define the relative temporal variation (RTV) of an internal state as the mean squared difference between the state and its flow-warped predecessor, divided by the state's spatial variance.
For raw inputs, RTV grows along inversion from 0.48 at the VAE latent to 1.09 at the endpoint.
Fusion removes 52--81\% of it in every latent, attention-bank, and VAE-skip state we measured, on all six clips (Supp.~\cref{tab:supp-rtv}).

\paragraph{The offset tolerates flow errors.}
A flow error of one latent pixel, eight image pixels, changes the transported offset by 0.10 in root mean square (RMS) and a transported state by 1.06.
Where flow is reliable, starting from the state differs from \cref{eq:start} by exactly the content mismatch $c_i=W_{k\to i}(\zinv_k(t_s))-\zinv_i(t_s)$.
This mismatch grows with the distance to the key frame, as does the LPIPS gap between the two variants, from 0.023 to 0.060 (Supp.~\cref{tab:supp-mismatch}).

\paragraph{A linear-Gaussian model.}
Both effects follow from a simple model (Supp.~\cref{sec:supp-theory}).
Assume the latent prior is Gaussian with variance $\lambda_k$ along direction $k$.
The ordinary differential equation that DDIM discretizes then multiplies direction $k$ by
\begin{equation}
  g_k(t)=\sqrt{\bar\alpha_t+\sigma_t^2/\lambda_k}\,,
  \label{eq:gain}
\end{equation}
where $\bar\alpha_t$ is the cumulative signal coefficient at time $t$ and $\sigma_t^2=1-\bar\alpha_t$.
Coarse, high-variance content is scaled by about $\sqrt{\bar\alpha_t}$, whereas low-variance directions, where per-frame noise lives, are raised to the noise level.
This is the amplification; fusion removes the noise before it happens.
The enhancement enters as AdaIN at the endpoint $T$.
Carried back to $t_s$, \cref{eq:gain} amplifies a constant shift 7.7 times but the finest scales only 0.6 times.
Indeed 97\% of the offset's energy lies in its channel means, which no flow error can move.

On the same clips, from the two ends of inversion alone, the model predicts the variance, temporal energy, and band power of all 14 interior inversion states with median errors of 2.3--2.4\%.
The offset's channel means are 6.75 times AdaIN's shift, below the predicted bound of 7.73.

\subsection{Runtime}
\label{sec:runtime}

On the six design clips of \cref{tab:zvrd} at $1280{\times}720$, timed in three processes with rotated method order, the pipeline takes 1.54\,s per frame, against 1.69\,s for E8 and 2.44\,s for E15 (Supp.~\cref{tab:supp-runtime}).
Key frames take 2.47\,s.
A non-key frame needs only 10 U-Net evaluations but still 1.43\,s, of which VAE encoding and decoding take about 0.68\,s.

\subsection{Transfer, long sequences, and stress tests}
\label{sec:transfer}

\begin{table}[t]
  \centering
  \setlength{\tabcolsep}{4pt}
  \fittable{%
  \begin{tabular}{@{}llccccc@{}}
    \toprule
    Set & Method & PSNR\up & SSIM\up & LPIPS\dn & Warp\dn & Flicker\dn \\
    \midrule
    SDE & MR E15 & 16.44 & 0.530\ns & 0.353 & 2.52 & 0.202 \\
     & MR E8 & 15.98 & 0.537\ns & \textbf{0.344} & 2.27 & 0.185 \\
     & MR state reuse & 14.85 & 0.481 & 0.416 & 3.76 & 0.278 \\
     & Fusion only & 16.01 & 0.537\ns & 0.371\ns & \textbf{1.50} & \textbf{0.153} \\
     & Late start only & 16.55 & 0.536\ns & 0.351 & 2.46 & 0.199 \\
     & Ours, $w_{\max}{=}0.5$ & \textbf{16.61}\ns & \textbf{0.541}\ns & 0.353 & 2.14 & 0.188 \\
     & Ours & \textbf{16.61} & 0.537 & 0.372 & 1.61 & 0.160 \\
    \midrule
    Long & MR E15 & 16.11 & 0.484 & \textbf{0.347}\ns & 1.32 & 0.167 \\
     & MR E8 & 15.82 & 0.480 & 0.390 & 1.13 & 0.161 \\
     & MR state reuse & 15.82 & 0.480 & 0.398 & 1.25 & 0.169 \\
     & Ours & \textbf{16.53} & \textbf{0.530} & 0.348 & \textbf{0.58} & \textbf{0.145} \\
    \midrule
    Stride 2 & MR E8 & 15.52 & 0.483 & 0.384 & 1.24 & 0.172 \\
     & Ours & \textbf{16.29} & \textbf{0.534} & \textbf{0.354} & \textbf{0.71} & \textbf{0.158} \\
    \midrule
    Stride 4 & MR E8 & 15.59 & 0.479 & 0.385\ns & 1.40 & 0.192 \\
     & Ours & \textbf{16.36} & \textbf{0.528} & \textbf{0.369} & \textbf{0.91} & \textbf{0.180} \\
    \midrule
    Scene cut & MR E8 & 17.33 & 0.693 & 0.254 & 1.56 & 0.143 \\
     & Ours & \textbf{17.83} & \textbf{0.726} & \textbf{0.223} & \textbf{1.13} & \textbf{0.130} \\
    \bottomrule
  \end{tabular}}
  \caption{\textbf{Transfer and stress tests.}
  SDE: 15 scenes of 30 frames at $346{\times}260$.
  Long: 300 frames at $1280{\times}720$.
  Stride 2 and 4: the long windows subsampled, 30 frames each.
  Long: four SDSD scenes outside the design and held-out splits. Scene cut: six sequences built from held-out clips, with a cut between frames 4 and 5; temporal metrics include the cut transition.
  All rows include exposure calibration. Fusion only: $\wmax=0.8$ on E8. Late start only: \cref{sec:latestart} without fusion. MR state reuse: \cref{sec:baselines}.
  Warp is in units of $10^{-2}$.
  Bold: best in column within each set. $^{\circ}$: the 95\% paired-bootstrap interval of the difference from our pipeline includes zero.}
  \label{tab:transfer}
\end{table}

\paragraph{Low resolution.}
The perceptual gain does not transfer to SDE (\cref{tab:transfer}).
The pipeline raises PSNR by 0.63\,dB (interval 0.43--0.83; on 14 of 15 scenes) and lowers warping error by 29\% and flicker by 13\%.
Its LPIPS, however, is worse than that of E8 by 0.028 (0.015--0.042), on 13 of 15 scenes.
The single-component rows place most of the loss in fusion: fusion alone costs 0.027 and late start alone 0.007.
The same holds at $256{\times}256$, where fusion alone costs 0.022 (Supp.~\cref{tab:supp-256}).
The fusion gate does not respond to resolution: the mean applied weight stays between 0.71 and 0.77 on every set (Supp.~\cref{tab:supp-fusion}).
Yet a $256{\times}256$ input carries about half as much frame-to-frame variation as a $1280{\times}720$ one (RTV 0.005 against 0.011), so the same averaging plausibly removes more detail relative to noise.
Lowering the ceiling to $\wmax=0.5$ reduces the LPIPS loss on SDE to 0.010 but gives up most of the temporal gain.

\paragraph{Long sequences.}
On the 300-frame windows the pipeline improves on E8 by 0.71\,dB in PSNR and 0.042 in LPIPS, on all four scenes, and halves warping error.
The gain does not decay with length: over ten consecutive 30-frame bins, the PSNR gain stays between 0.62 and 0.77\,dB.

\paragraph{Faster motion and scene cuts.}
At stride 2 and 4 the pipeline gains 0.77\,dB in PSNR in both cases and lowers warping error by 43\% and 35\%.
The LPIPS gain shrinks from 0.030 to 0.016, and at stride 4 it is within the interval.
A scene cut costs the pipeline the rest of its key interval (\cref{fig:temporal}c).
It is briefly worse than E8 right after the cut, and the next key frame restores the full advantage.
Over whole sequences it remains better than E8 (0.50\,dB, 0.031 LPIPS).

\section{Limitations}
\label{sec:limitations}

With $\wmax=0.8$ the perceptual gain does not transfer below $1280{\times}720$: LPIPS is worse than that of E8 by 0.028 on SDE and by 0.015 at $256{\times}256$.
All constants were chosen on clips from the same dataset and capture rig as the held-out clips, and the ZVRD comparison uses six of the design clips.
SDSD and SDE are robot-captured with slow motion, and fast motion and scene cuts are tested only with sets derived from SDSD.
Each key-frame refresh changes the output more than a non-key transition, and a scene cut costs the rest of its key interval.
ZVRD is compared only through our implementation, as it has no public code.
The trained Zero-TIG has higher PSNR and SSIM and is about $12\times$ faster, and against the longer MR schedules the LPIPS gain is not statistically resolved on 13 clips; on SDE and at $256{\times}256$, E15 has lower LPIPS than our pipeline.
PSNR and flow-based metrics can reward smoothing; LPIPS guards against this only in part, and we ran no user study.

\section{Conclusion}
\label{sec:conclusion}

Training-free diffusion enhancement of low-light video does not need hundreds of sampling steps or test-time optimization: eight-step framewise processing with a frozen image prior is already over two orders of magnitude faster than ZVRD and scores better.
Input fusion and late-start offset propagation make it faster and better at $1280{\times}720$, on held-out clips and on 300-frame sequences.

{
    \small
    \bibliographystyle{ieeenat_fullname}
    \bibliography{main}
}

\clearpage
\setcounter{page}{1}
\maketitlesupplementary

\section{Implementation details}
\label{sec:supp-impl}

\paragraph{Backbone.}
We use the released MR implementation~\cite{cho2026mr} with Stable Diffusion v1.5, a float16 U-Net, a float32 VAE, and empty text conditioning.
The latent is the posterior mean times 0.18215.
AdaIN normalizes the inversion endpoint to style noise drawn once with a fixed seed.
The two framewise schedules are
\begin{align*}
  \mathcal T_{15} &= \{1,67,133,199,265,331,397,463,\\
                  &\qquad 529,595,661,727,793,859,925\},\\
  \mathcal T_{8}  &= \{1,126,251,376,501,626,751,876\}.
\end{align*}
When every token of a self-attention layer is replaced by the injected bank, we skip the layer's own attention computation; this leaves the output unchanged.
SDE frames ($346{\times}260$) are encoded at $352{\times}264$ and decoded back to the original size.

\paragraph{Flow for fusion.}
Farneb\"ack flow~\cite{farneback2003} is estimated on grayscale frames downsampled so that the long side is at most 320 pixels (pyramid scale 0.5, three levels, window 21, three iterations).
A location is reliable if its warp coordinate is in bounds and its forward-backward error is at most $1.5+0.05\lVert b\rVert$ pixels, where $b$ is the backward displacement.
The flow is resized bilinearly to the working resolution, with vectors rescaled per axis; the mask is resized by nearest neighbor.

\paragraph{Latent transport.}
Flow from the key frame's fused input to the current fused input is estimated in the same way and resized to the latent grid ($160{\times}90$ for $1280{\times}720$).
A latent location is reliable, $M_i=1$, only if at least 99.9\% of its image footprint is reliable and its warp coordinate is in bounds.

\paragraph{State-reuse baseline.}
At non-key frames this baseline warps the key frame's inversion endpoint and attention bank, falls back to current-frame noise and attention where flow is unreliable, and runs all 15 denoising steps.
With $K=10$ it needs $(3\cdot30+27\cdot15)/30=16.5$ U-Net evaluations per frame on a 30-frame clip.

\paragraph{Algorithm.}
For each frame $i=0,1,\dots$:
\begin{enumerate}
  \item Compute the fused input $f_i$ with \cref{eq:fusion,eq:history,eq:weight,eq:gate} and encode it, keeping the VAE skips.
  \item If $i\equiv0\pmod K$: invert over $\mathcal T_{15}$ and record the attention bank, apply AdaIN, denoise over $\mathcal T_{15}$, and store $f_i$, $\Delta_k$ of \cref{eq:offset}, and $\bar\Delta_k$.
  \item Otherwise: invert over $\mathcal T_s$ and record the attention bank, estimate flow from the stored key input to $f_i$, form the starting state with \cref{eq:start}, and denoise over $\mathcal T_s$.
  \item Decode with the current skips to obtain $y_i$, and calibrate $y_i$ against $x_i$ to obtain $\tilde y_i$.
\end{enumerate}

\section{Exposure calibration}
\label{sec:supp-calib}

The calibration applies one scalar gain to the three linear RGB channels.
Luminance statistics are computed on linear Rec.\,709 luminance at a long side of at most 320 pixels.
Let $\mu_x$ and $\mu_y$ be the input and output means and $q_c$, $q_h$ the 0.995 and 0.999 quantiles of the input.
With $\varepsilon=10^{-6}$,
\begin{align}
  m_i &= 0.9\,\mu_x/q_c, \\
  r_i &= \tfrac12\log_2\frac{\max(m_i,\varepsilon)}{\max(\mu_y,\varepsilon)}, \\
  b_i &= \operatorname{clip}(r_i,-0.5,0.5), \\
  L_i &= 0.9\,\mu_x/\max(q_h,\varepsilon).
\end{align}
The guard holds when $L_i\le\mu_y\le m_i$: the output is darker than the white point $q_c$ asks for, but a brighter white point $q_h$ would no longer ask for brightening.
The guarded request is $\hat e_i=0$ if the guard holds and $\hat e_i=b_i$ otherwise.
The state starts at $e_0=\hat e_0$, and for $i>0$
\begin{equation}
  e_i=0.8\,e'_{i-1}+0.2\,\hat e_i,
\end{equation}
where $e'_{i-1}=0$ if the guard holds and $e_{i-1}>0$, or if the guard does not hold and $e_{i-1}r_i<0$; otherwise $e'_{i-1}=e_{i-1}$.
We apply $2^{e_i}$ in linear RGB, clip to $[0,1]$, and convert to sRGB.
The state is reset only at the start of a sequence.
The heuristic can fail, for example by darkening an output that is already dark.

\section{ZVRD implementation}
\label{sec:supp-zvrd}

ZVRD~\cite{cao2025zvrd} has no official implementation; at the time of writing the repository linked by its authors contains only a README.
We implemented it from the paper, on the Generative Diffusion Prior code~\cite{fei2023gdp} that its low-light experiments use, with its 950-step sampling configuration and RAFT flow~\cite{teed2020raft}.
Settings that the paper leaves open were fixed without access to ground truth.
At $256{\times}256$ each frame is one full-frame patch, which we verified to give the same output as the tiled code path at this size.
At $1280{\times}720$ the patch strategy tiles each frame into 50 overlapping $256{\times}256$ patches at every step.
Because of the cost, the $1280{\times}720$ run covers one clip of 30 frames, the one in \cref{fig:teaser}, and is used only for the time shown there.
All ZVRD and MR runs use the same NVIDIA L40.

\section{Data splits}
\label{sec:supp-data}

\begin{table}[t]
  \centering
  \small
  \setlength{\tabcolsep}{4pt}
  \fittable{%
  \begin{tabular}{@{}p{0.22\linewidth}p{0.36\linewidth}p{0.34\linewidth}@{}}
    \toprule
    Split & SDSD scenes & Used to choose \\
    \midrule
    Design (12) & indoor 13, 15, 21, 23, 31, 33; outdoor 1, 5, 14, 36, 46, 48 & $\wmax$, $t_s$, $K$; calibration \\
    Held-out (6) & indoor 50, 52, 58; outdoor 49, 60, 62 & calibration \\
    Held-out (7) & indoor 60, 68, 70; outdoor 63, 66, 75, 76 & neither \\
    \bottomrule
  \end{tabular}}
  \caption{\textbf{Roles of the 25 SDSD clips}: which clips were used to choose the settings of the temporal components and of the exposure calibration. The temporal components never ran on a held-out clip before their settings were fixed. The E8 schedule was chosen from framewise measurements on the SDSD test scenes.}
  \label{tab:supp-splits}
\end{table}

\Cref{tab:supp-splits} lists the clips.
On the design clips we compared $\wmax\in\{0.5,0.8\}$ and $t_s\in\{265,397\}$ and chose $\wmax=0.8$ and $t_s=265$.
The key interval $K=10$ is that of the state-reuse baseline, for which $K\in\{5,10,20\}$ were compared on six design clips.
The gate constants of \cref{eq:weight,eq:gate} were set by hand and not varied.
The long windows are outdoor scenes 2, 19, 22, and 37, which are not in the SDSD test configuration and were used for no setting.
Each scene-cut sequence joins two held-out clips of the same indoor or outdoor group.
SDE was used for no setting.

\section{Evaluation and timing details}
\label{sec:supp-eval}

SSIM uses an $11{\times}11$ Gaussian window with standard deviation 1.5 and is averaged over RGB channels.
Evaluation flow is bidirectional Farneb\"ack flow on the grayscale references at a long side of at most 640 pixels.
A pixel is valid if its warp coordinate is in bounds and its forward-backward error is at most $1.5+0.01(\lVert b\rVert+\lVert f\rVert)$ pixels, with $b$ and $f$ the backward and forward flows.
The same mask is used for every method.
For flicker, let $p_i$ be the output and $g_i$ the reference, $Y$ linear Rec.\,709 luminance, and $a_i=\ln\max(Y(p_i),10^{-4})-\ln\max(Y(g_i),10^{-4})$.
Flicker is the mean of $|a_i-W_{i-1\to i}(a_{i-1})|$ over valid pixels.
The reference flow is an estimate, not ground-truth motion, and is never an input to inference.

Timing starts when the low-light image is decoded and ends when the calibrated output is on the CPU.
Model loading, warm-up, and file writing are excluded.
The timings of E8, E15, and our pipeline ran on a GPU with no other process.
Bootstrap intervals are descriptive and are not adjusted for multiple comparisons.

\begin{table}[t]
  \centering
  \small
  \setlength{\tabcolsep}{4pt}
  \fittable{%
  \begin{tabular}{@{}lcccc@{}}
    \toprule
    Method & s/frame\dn & P95 (s) & NFE & Memory (GiB) \\
    \midrule
    MR E8  & 1.69 & \textbf{1.70} & 16 & \textbf{6.1} \\
    MR E15 & 2.44 & 2.46 & 30 & 6.4 \\
    Ours   & \textbf{1.54} & 2.49 & \textbf{12} & 6.4 \\
    \bottomrule
  \end{tabular}}
  \caption{\textbf{Repeated timing at $1280{\times}720$}: the six design clips of \cref{tab:zvrd}, 30 frames each, three processes with rotated method order on one NVIDIA L40. NFE is the number of network evaluations per frame. P95 is the 95th-percentile frame time. Memory is the peak allocated GPU memory. The three process means differ by at most 0.013\,s for each method. Bold: best in column.}
  \label{tab:supp-runtime}
\end{table}

\section{Additional results}
\label{sec:supp-results}

\begin{table}[t]
  \centering
  \setlength{\tabcolsep}{3pt}
  \fittable{%
  \begin{tabular}{@{}lccccccr@{}}
    \toprule
    Method & PSNR\up & SSIM\up & LPIPS\dn & Warp\dn & Flicker\dn & NFE & s/frame\dn \\
    \midrule
    MR E8 & 18.91\ns & 0.672 & 0.163 & 0.94 & 0.103 & 16 & 0.38 \\
    Fusion only & 19.10\ns & 0.683\ns & 0.186 & \textbf{0.50} & \textbf{0.086} & 16 & 0.41 \\
    Late start only & 18.97 & 0.678 & \textbf{0.156} & 0.96 & 0.105 & \textbf{12} & \textbf{0.32} \\
    Ours, $w_{\max}{=}0.5$ & 19.08 & \textbf{0.688}\ns & \textbf{0.156} & 0.65 & 0.091 & \textbf{12} & 0.35 \\
    Ours & \textbf{19.18} & \textbf{0.688} & 0.178 & 0.51 & 0.087 & \textbf{12} & 0.35 \\
    \bottomrule
  \end{tabular}}
  \caption{\textbf{Single components at $256{\times}256$}, the six design clips of \cref{tab:zvrd}. Fusion only: $\wmax=0.8$ on E8; late start only: no fusion. Warp is in units of $10^{-2}$; NFE is the number of network evaluations per frame. Bold: best in column. $^{\circ}$: the 95\% paired-bootstrap interval of the difference from our pipeline includes zero.}
  \label{tab:supp-256}
\end{table}

\begin{table}[t]
  \centering
  \small
  \setlength{\tabcolsep}{4pt}
  \fittable{%
  \begin{tabular}{@{}lcccc@{}}
    \toprule
    Set & Clips & $\hat\sigma_i$ & Mean $w_i$ & Reliable \\
    \midrule
    SDSD, $1280{\times}720$, 30 frames  & 25 & 0.0055 & 0.755 & 0.992 \\
    SDSD, $1280{\times}720$, 300 frames$^{\ast}$ & 6  & 0.0057 & 0.760 & 0.992 \\
    SDSD, $256{\times}256$              & 6  & 0.0059 & 0.770 & 0.989 \\
    SDE, $346{\times}260$               & 15 & 0.0022 & 0.708 & 0.990 \\
    \bottomrule
  \end{tabular}}
  \caption{\textbf{Fusion statistics of our pipeline}, averaged over non-initial frames and then clips: the scale $\hat\sigma_i$ of \cref{eq:weight}, the applied weight, and the fraction of pixels with reliable flow. The SDSD rows include design clips. $^{\ast}$The four long scenes and two design scenes.}
  \label{tab:supp-fusion}
\end{table}

\Cref{tab:supp-256} gives the single-component rows at $256{\times}256$.
Fusion alone raises LPIPS relative to E8 by 0.022, on all six clips, and late start alone lowers it by 0.007.
\Cref{tab:supp-fusion} shows that the fusion gate rarely closes on these slow-motion data, so fusion acts almost as plain motion-compensated recursive averaging, and that the applied weight is nearly the same at every resolution.

\section{Measurements inside the model}
\label{sec:supp-mechanism}

\begin{table}[t]
  \centering
  \setlength{\tabcolsep}{3pt}
  \fittable{%
  \begin{tabular}{@{}lcccccc@{}}
    \toprule
    & \multicolumn{3}{c}{$1280{\times}720$} & \multicolumn{3}{c}{$256{\times}256$} \\
    \cmidrule(lr){2-4}\cmidrule(l){5-7}
    State & Raw & Fused & Reduction & Raw & Fused & Reduction \\
    \midrule
    Input luminance               & 0.011 & 0.002 & 78\% & 0.005 & 0.001 & 74\% \\
    VAE latent $z_0$              & 0.477 & 0.171 & 65\% & 0.168 & 0.066 & 59\% \\
    Inversion latent, $t=67$      & 0.607 & 0.234 & 62\% & 0.222 & 0.084 & 62\% \\
    Inversion latent, $t_s=265$   & 0.852 & 0.364 & 58\% & 0.355 & 0.134 & 64\% \\
    Inversion endpoint, $t=925$   & 1.092 & 0.506 & 55\% & 0.491 & 0.192 & 63\% \\
    VAE skip, level 0             & 0.016 & 0.003 & 81\% & 0.009 & 0.002 & 78\% \\
    VAE skip, level 1             & 0.589 & 0.125 & 79\% & 0.171 & 0.036 & 78\% \\
    VAE skip, level 2             & 0.669 & 0.172 & 74\% & 0.221 & 0.045 & 79\% \\
    VAE skip, level 3             & 0.642 & 0.190 & 71\% & 0.201 & 0.047 & 77\% \\
    Attention bank (range) & 0.08--0.29 & 0.03--0.11 & 52--62\% & 0.19--0.39 & 0.09--0.17 & 46--64\% \\
    \bottomrule
  \end{tabular}}
  \caption{\textbf{Relative temporal variation (RTV) of internal states} for raw and fused input, six design clips of 30 frames. The attention row gives the range over three latent resolutions and three times, $t\in\{1,265,925\}$. Reduction is $1-\mathrm{RTV}_{\text{fused}}/\mathrm{RTV}_{\text{raw}}$ per clip, averaged; it is positive on every clip in every row.}
  \label{tab:supp-rtv}
\end{table}

\begin{table}[t]
  \centering
  \small
  \setlength{\tabcolsep}{5pt}
  \fittable{%
  \begin{tabular}{@{}cccc@{}}
    \toprule
    Key age & Mismatch / offset & Mismatch / content & LPIPS gap \\
    \midrule
    1 & 0.69 & 0.59 & 0.023 \\
    2 & 0.85 & 0.72 & 0.037 \\
    3 & 0.94 & 0.80 & 0.043 \\
    4 & 1.00 & 0.84 & 0.049 \\
    5 & 1.03 & 0.87 & 0.050 \\
    6 & 1.06 & 0.89 & 0.051 \\
    7 & 1.08 & 0.92 & 0.056 \\
    8 & 1.10 & 0.93 & 0.059 \\
    9 & 1.12 & 0.94 & 0.060 \\
    \bottomrule
  \end{tabular}}
  \caption{\textbf{Content mismatch of state transport by key age} (frames since the key frame), six design clips, 18 frames per age. Columns: RMS of $c_i$ relative to the RMS of the transported offset; RMS of $c_i$ relative to the spatial RMS of the current inversion latent; mean LPIPS of the key-state variant minus that of the offset variant.}
  \label{tab:supp-mismatch}
\end{table}

\paragraph{RTV.}
For a state $s$, RTV at frame $i$ is the masked mean of $|s_i-W(s_{i-1})|^2$ divided by the masked spatial variance of $s_i$, where $W$ warps with the reference flow of \cref{sec:supp-eval} resized to the state's grid.
Each frame of the six design clips was inverted over $\mathcal T_{15}$ twice, from the raw frame and from the fused input.
\Cref{tab:supp-rtv} reports the result.

\paragraph{Transport.}
On the 18 key frames of the same clips we recorded $\zinv_k(t_s)$ and $\Delta_k$.
After removing channel means, the spatial RMS of the offset is 0.108, against 0.871 for the state $\zden_k(t_s)$.
A displacement of one latent pixel changes the offset by 0.097 RMS and the state by 1.059.
\Cref{tab:supp-mismatch} reports the content mismatch $c_i$ by key age.

\section{A linear-Gaussian model of inversion}
\label{sec:supp-theory}

\paragraph{Assumptions.}
(A1)~Near the data, the latent prior is Gaussian with mean $\mu$ and covariance $U\Lambda U^\top$, $\Lambda=\mathrm{diag}(\lambda_k)$.
(A2)~A low-light frame is $x_i=s_i+n_i$, where $s_i$ follows the flow and $n_i$ is noise of variance $\nu$ per pixel, independent across frames and of $s_i$.
(A3)~The prior is stationary, so $U$ is the Fourier basis, and warps are translations, which preserve variance and commute with the gain of inversion.

\paragraph{Gain of inversion.}
Let $u_t$ be the coordinate of $z_t-\sqrt{\bar\alpha_t}\,\mu$ along direction $k$, and write $\lambda$ for $\lambda_k$.
The forward process gives $u_t=\sqrt{\bar\alpha_t}\,u_0+\sigma_t\epsilon$ with $\sigma_t^2=1-\bar\alpha_t$.
Then
\begin{align}
  \mathrm{Var}(u_t) &= \bar\alpha_t\lambda+\sigma_t^2
     && \text{(A1)} \label{eq:s-var}\\
  \nabla\log p_t(u_t) &= -\frac{u_t}{\bar\alpha_t\lambda+\sigma_t^2}
     && \text{Gaussian score} \label{eq:s-score}\\
  \epsilon^\star(u_t) &= \frac{\sigma_t\,u_t}{\bar\alpha_t\lambda+\sigma_t^2}
     && \epsilon^\star=-\sigma_t\nabla\log p_t \label{eq:s-eps}
\end{align}
Deterministic DDIM discretizes $d\bar u/d\tau=\epsilon^\star$ in the variables $\bar u=u_t/\sqrt{\bar\alpha_t}$ and $\tau=\sigma_t/\sqrt{\bar\alpha_t}$~\cite{song2022ddim}.
Substituting $u_t=\sqrt{\bar\alpha_t}\,\bar u$, with $\bar u(0)=u_0$,
\begin{align}
  \frac{d\bar u}{d\tau} &= \frac{\sigma_t\sqrt{\bar\alpha_t}\,\bar u}{\bar\alpha_t\lambda+\sigma_t^2}
     && \text{\eqref{eq:s-eps}} \\
   &= \frac{\tau\,\bar u}{\lambda+\tau^2}
     && \div\,\bar\alpha_t \label{eq:s-ode}\\
  \ln\frac{\bar u(\tau)}{\bar u(0)} &= \tfrac12\ln\frac{\lambda+\tau^2}{\lambda}
     && \text{integrate} \\
  \bar u(\tau) &= \bar u(0)\sqrt{1+\tau^2/\lambda}
     && \text{exponentiate} \\
  u_t &= u_0\sqrt{\bar\alpha_t+\sigma_t^2/\lambda}
     && \times\sqrt{\bar\alpha_t},\ \bar\alpha_t\tau^2=\sigma_t^2 \label{eq:s-gain}
\end{align}
\Cref{eq:s-gain} is \cref{eq:gain} of the main paper.

\paragraph{Affine law.}
Let $Q_t$ be a quadratic statistic that is diagonal in $U$, with per-direction value $q_k$ at $t=0$, and let $Q_\infty=\sum_k q_k/\lambda_k$.
\begin{align}
  Q_t &= \textstyle\sum_k g_k(t)^2\,q_k
     && \text{\eqref{eq:s-gain}} \\
      &= \bar\alpha_t\textstyle\sum_k q_k+\sigma_t^2\sum_k q_k/\lambda_k
     && \text{\eqref{eq:gain}} \\
      &= \bar\alpha_t\,Q_0+\sigma_t^2\,Q_\infty
     && \text{collect} \label{eq:s-affine}
\end{align}
Under (A3), spatial variance, temporal-difference energy, lag-one autocovariance, and band power are such statistics and obey \cref{eq:s-affine}.
RTV is a ratio of two such statistics.
As $t$ grows, the weight $\sigma_t^2/\lambda_k$ favors low-variance directions, so RTV rises whenever frame-to-frame change is relatively larger there, as it is for noise.
This is the amplification in \cref{tab:supp-rtv}.

\paragraph{Fusion as exponential smoothing.}
With a constant weight $w$ in \cref{eq:fusion}, (A2), and (A3), the residual noise $\eta_i$ of the fused input obeys the following along a motion trajectory, where $v$ is its stationary variance and $\delta_i=\eta_i-\eta_{i-1}$.
\begin{align}
  \eta_i &= (1-w)\,n_i+w\,\eta_{i-1}
     && \text{\eqref{eq:fusion}, (A3)} \\
  v &= (1-w)^2\nu+w^2v
     && n_i\perp\eta_{i-1} \\
  v &= \nu\,\frac{1-w}{1+w}
     && \text{solve} \label{eq:s-v}\\
  \mathrm{E}\,\delta_i^2 &= 2v(1-w)
     && \text{lag-one corr.\ } w \\
    &= 2\nu\,\frac{(1-w)^2}{1+w}
     && \text{\eqref{eq:s-v}} \label{eq:s-gamma}
\end{align}
Raw input has frame-difference noise energy $2\nu$.
At the mean weight measured on the six design clips, $w=0.754$, \cref{eq:s-gamma} leaves about 3.5\% of it; this is approximate, because the weight varies over pixels.

\paragraph{The offset is mostly a constant.}
The enhancement enters as AdaIN at $T=925$, whose shift is constant in space.
Carried back to $t_s$, a perturbation along direction $k$ is multiplied by $G_k=g_k(t_s)/g_k(T)$.
For the constant direction ($\lambda_k\to\infty$), $G_k\to\sqrt{\bar\alpha_{t_s}/\bar\alpha_T}=7.73$; for the finest scales ($\lambda_k\to0$), $G_k\to\sigma_{t_s}/\sigma_T=0.60$.
Denoising thus acts on the enhancement as a low-pass filter.
On the 18 key frames, 97.4\% of the offset's energy is in its channel means.

\paragraph{Testing the model.}
\Cref{tab:supp-affine} tests \cref{eq:s-affine}: from the VAE latent and the endpoint $t=925$ alone, we predict four statistics at the 14 interior times of $\mathcal T_{15}$.
The largest errors are an underestimate of temporal energy by 12--15\% at $t=67$--$199$ at $1280{\times}720$: the first inversion steps amplify frame-to-frame variation somewhat more than the model does.
Regressing the offset's channel means on AdaIN's channel shifts through the origin gives a slope of 6.75 (correlation 0.997, 72 values), below the bound of 7.73.

The model treats noise propagation as linear.
\Cref{tab:supp-noise} shows that this fails: doubling injected noise multiplies the response energy by 4.00 only in the first VAE convolution, and by 1.75--2.77 in every later state.
The model therefore explains where noise is amplified and why the offset transports well, but not how much noise each state receives.

\begin{table}[t]
  \centering
  \setlength{\tabcolsep}{3pt}
  \fittable{%
  \begin{tabular}{@{}llccc@{}}
    \toprule
    Resolution & Input & $V$ and $N$ (median / max) & Band power (median / max) & $\rho_1$ (median / max) \\
    \midrule
    $1280{\times}720$ & Raw   & 4.2\% / 14.0\% & 3.5\% / 11.5\% & 0.001 / 0.027 \\
    $1280{\times}720$ & Fused & 3.5\% / 14.6\% & 3.3\% / 11.5\% & 0.002 / 0.027 \\
    $256{\times}256$  & Raw   & 1.6\% / 8.3\%  & 1.7\% / 4.1\%  & 0.001 / 0.008 \\
    $256{\times}256$  & Fused & 2.1\% / 5.4\%  & 2.1\% / 4.8\%  & 0.001 / 0.006 \\
    \bottomrule
  \end{tabular}}
  \caption{\textbf{Errors of the affine law} (\cref{eq:s-affine}) at the 14 interior times of $\mathcal T_{15}$, predicted from the two ends of inversion, six design clips. $V$: spatial variance; $N$: temporal-difference energy; band power: eight radial frequency bands; $\rho_1$: lag-one spatial autocorrelation. Errors are relative for $V$, $N$, and band power and absolute for $\rho_1$. Over all cells the median error is 2.4\% for $V$ and $N$ and 2.3\% for band power.}
  \label{tab:supp-affine}
\end{table}

\begin{table}[t]
  \centering
  \small
  \setlength{\tabcolsep}{5pt}
  \fittable{%
  \begin{tabular}{@{}lcc@{}}
    \toprule
    State & Energy ratio & Relative response \\
    \midrule
    VAE skip, level 0 (first convolution) & 4.00 & 0.005 \\
    VAE skip, levels 1--3 & 2.20--2.49 & 0.14--0.18 \\
    VAE latent $z_0$ & 2.21 & 0.10 \\
    Inversion latent, $t=67$ & 2.11 & 0.13 \\
    Inversion latent, $t_s=265$ & 1.88 & 0.21 \\
    Inversion endpoint, $t=925$ & 1.75 & 0.42 \\
    Attention bank (range) & 2.19--2.77 & 0.06--0.14 \\
    \bottomrule
  \end{tabular}}
  \caption{\textbf{Response to injected noise}, 60 fused frames of the six design clips at $1280{\times}720$. White noise $\xi$ with standard deviation 0.01 is added to the input. The attention row gives the range over three latent resolutions and three times. Energy ratio: median over frames of the response energy for $2\xi$ over that for $\xi$; a linear map gives 4. Relative response: median response energy for $\xi$ divided by the state's variance.}
  \label{tab:supp-noise}
\end{table}

\end{document}